\documentclass[runningheads]{llncs}
\usepackage[T1]{fontenc}
\usepackage{graphicx}
\usepackage{amsmath}
\usepackage{amssymb}
\usepackage{multirow}
\usepackage{booktabs}
\usepackage{siunitx}
\usepackage{graphicx}
\usepackage{array}

\newcolumntype{N}{>{\centering\arraybackslash}m{.7in}}

\newcolumntype{M}{>{\raggedright\arraybackslash}m{.7in}}

\begin{document}
\title{Beyond Pixels: Leveraging the Language of Soccer to Improve Spatio-Temporal Action Detection in Broadcast Videos}
%
\titlerunning{Beyond Pixels: Leveraging the Language of Soccer}
%
\author{Jeremie Ochin\inst{1,2}\orcidID{0009-0008-3909-3757} \and
Raphael Chekroun\inst{2} \and
Bogdan Stanciulescu\inst{1} \and
Sotiris Manitsaris\inst{1}\orcidID{0000-0003-4552-1793}}
\authorrunning{J. Ochin et al.}
%
\institute{Centre for Robotics, Mines Paris - PSL, France \and
Footovision, Paris, France\\
\email{\{jeremie.ochin,bogdan.stanciulescu,sotiris.manistsaris\}@minesparis.psl.eu}, raphael.chekroun@footovision.com}
\maketitle              
\begin{abstract}
State-of-the-art spatio-temporal action detection (STAD) \linebreak methods show promising results for extracting soccer events from broadcast videos. However, when operated in the high-recall, low-precision regime required for exhaustive event coverage in soccer analytics, their lack of contextual understanding becomes apparent: many false positives could be resolved by considering a broader sequence of actions and game-state information. In this work, we address this limitation by reasoning at the game level and improving STAD through the addition of a denoising sequence transduction task. Sequences of noisy, context-free player-centric predictions are processed alongside clean game state information using a Transformer-based encoder-decoder model. By modeling extended temporal context and reasoning jointly over team-level dynamics, our method leverages the "language of soccer"—its tactical regularities and inter-player dependencies—to generate "denoised" sequences of actions. This approach improves both precision and recall in low-confidence regimes, enabling more reliable event extraction from broadcast video and complementing existing pixel-based methods.

\end{abstract}
\section{Introduction}

Fully automating the production of high-quality soccer analytics from broadcast video remains a challenge. While the past decade has brought major advances in computer vision and machine learning—enabling reliable player tracking, identification, and pitch localization—fine-grained event annotation remains largely a manual, time-consuming process carried out by expert operators who must scrub through footage to accurately label actions \cite{ochin2025}. Spatio-temporal action detection (STAD) methods, operated at low confidence thresholds to maximize recall, are well-suited to assist annotation operators by proposing candidate actions for validation.

\begin{figure}[h!]\centering
\includegraphics[width=.99\linewidth]{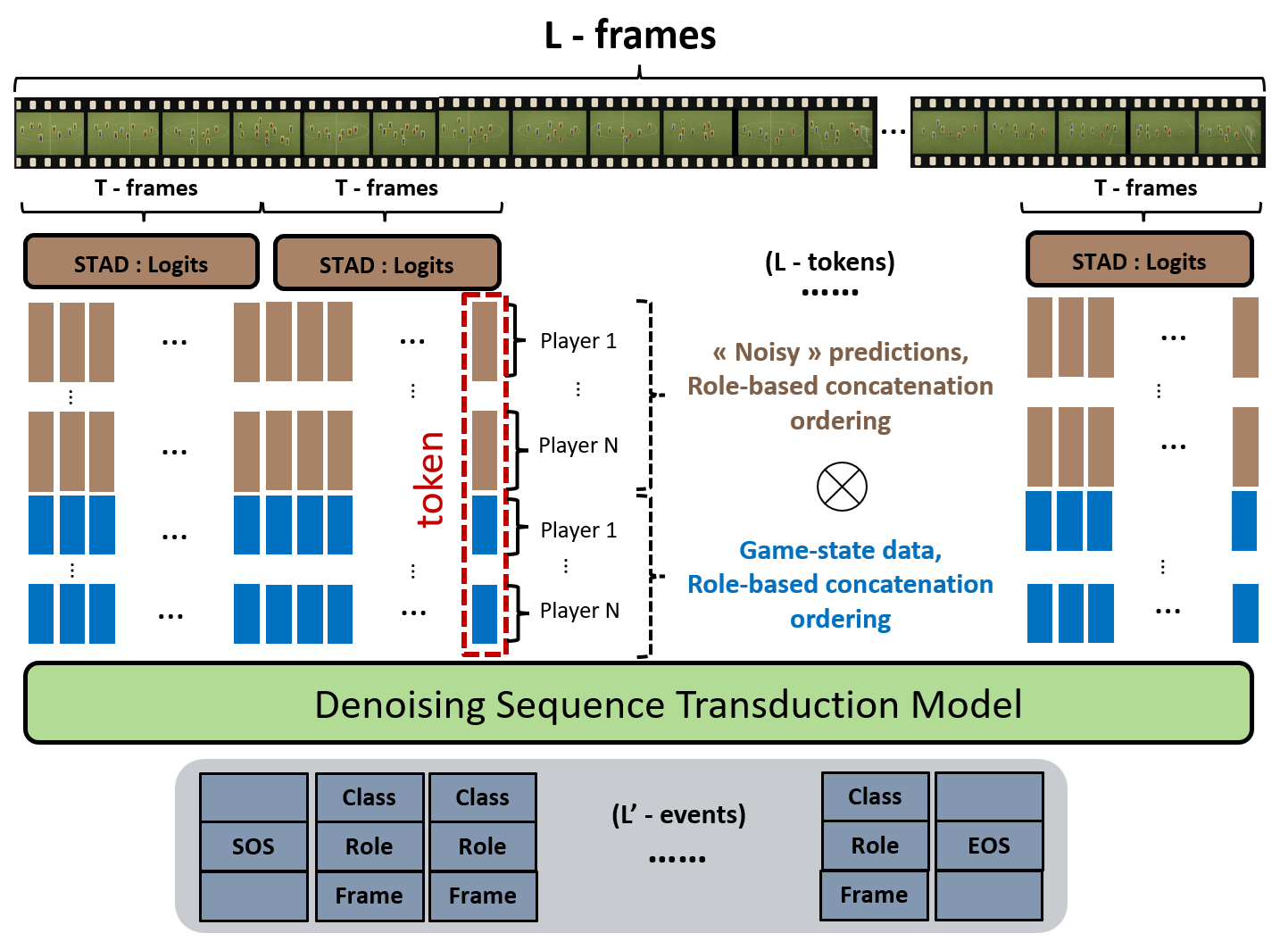}
\caption{Overview of our method. A STAD model first produces per-frame, per-player action predictions over a long sequence. These predictions are concatenated in a consistent role-based order, using metadata. Each token is constructed by combining these structured prior predictions with additional game-state features such as player positions and velocities. The resulting sequence of tokens is fed into an encoder-decoder Transformer, which is trained to auto-regressively generate cleaned action sequences—predicting, for each action, its class, associated player, and frame number. Our method emphasizes the importance of both temporal and inter-player context, leveraging a wide temporal window and structured game-state information.}
\label{fig:schema_global}
\end{figure}

STAD aims to identify \textbf{when} an action occurs, \textbf{what} it is, and \textbf{who} performed it—or \textbf{where}. In practice, it involves detecting and classifying player actions in untrimmed video, producing temporally grounded “action tubes” for events such as ball drives, passes, crosses, shots, headers, throw-ins, ball blocks, and tackles ("on-the-ball events").

Despite notable improvements in STAD accuracy \cite{STAR_survey}, key limitations remain. Some stem from the nature of the game and its recording conditions—occlusions, motion blur, rapid camera movements, and frequent visual ambiguities. Another frequent constraint is the limited computational budget available for training or inference, which often forces fine-grained action recognition models to process only short clips (16–64 frames). However, this approach overlooks the contextual richness of soccer, where the relevance and tactical significance of an action are strongly influenced by surrounding players and the current phase of play. This limitation is especially apparent in high-recall settings, where many false positives could be avoided with access to long-range temporal and game-state context.

To address these limitations, we augment existing STAD models with a "denoising" sequence transduction task. We process long sequences of context-free player-centric action predictions—alongside structured game-state data (e.g., player positions, velocities, team IDs)—using a transformer-based encoder-decoder (Figure~\ref{fig:schema_global}). Inspired by natural language processing pretraining strategies, our model learns to reconstruct "clean", coherent and tactically plausible action sequences from corrupted inputs, akin to translating noisy text into fluent language.

In doing so, our method bridges two previously separate strands of research: (i) image-based STAD and (ii) sequence modeling of game events. By modeling the "language of soccer"—capturing the tactical, temporal, and spatial dependencies between actions—our approach goes beyond conventional detection and remains agnostic to the underlying detection model. This enhances robustness in real-world settings, where visual noise and ambiguity are common.

We evaluate our approach using a dataset of 299 full-length games with broadcast footage, tracking data, and player metadata. Designed for real-world use, our model is trainable on a single consumer-grade or entry-level professional GPU.

This paper makes three main contributions. We first introduce a novel formulation of spatio-temporal action detection (STAD) as a denoising sequence transduction task, leveraging structured game-state context to guide the reconstruction of action sequences. We then propose a transformer-based model that jointly encodes noisy visual predictions and player-centric features to produce coherent, tactically plausible action sequences—enabling reasoning beyond raw pixel data. Finally, we provide empirical evidence that our approach improves detection performance in high-recall settings, while remaining computationally efficient.

This work offers a new perspective on soccer video analysis by demonstrating how integrating prior image-based predictions with structured context-rich representations of game dynamics can improve the accuracy and coherence of event detection.

\section{Related Work}
\subsection{Spatio-Temporal Action Detection (STAD)}

Numerous techniques have been developed for STAD in recent years. These approaches can be categorized into frame-level and clip-level models~\cite{Multisports2021}. The frame-level models predict the bounding box and action type for each frame and then integrate these predictions. Conversely, clip-level methods, also called action tubelet detectors, endeavour to model both temporal context and action localization. A comprehensive review is provided in the survey by Wang et al.~\cite{STAR_survey}.

Among the clip-level methods, Track-Aware Action Detector (TAAD)~\cite{singh2022} yields per-actor, per-frame action predictions by first detecting and tracking players, and then aggregating features along player trajectories using a fine-tuned 3D CNN and ROI Align~\cite{MaskR2018}, followed by a Temporal Convolutional Network. TAAD demonstrates state-of-the-art performance on public STAD benchmarks and exhibits robustness to camera motion.

This method is particularly well-suited for soccer analytics, where tracking, identification, and position estimation typically precede event annotation. With TAAD, actions are naturally linked to existing player tracklets and identities, which is important since soccer analytics often requires the construction of statistics on a per-player basis.

\subsection{Generative Models for Soccer Action Sequences}

Traditional approaches have relied heavily on Markov models \cite{markovPena,markovVanRoy}, which align well with soccer’s structure as a sequence of discrete, observable states, such as possession changes, spatial transitions and goals.

Recent work has drawn inspiration from generative models to better capture the sequential and contextual structure of soccer. Simpson et al. \cite{Seq2Event2022} used RNNs and transformers to jointly predict the location and type of the next action from 40 prior events, outperforming Markovian baselines. Mendes-Neves et al. \cite{MendesNeves2024} introduced a Large Event Model trained on the previous three actions to predict the next action’s position, type, and success, enabling downstream tasks such as tactical analysis and performance evaluation. Baron et al. \cite{Baron2024} framed next-action prediction as a classification task over tokenized events using a transformer decoder, improving temporal modeling over prior MLP-based approaches.

These models contribute to the advancement of generative modeling of soccer action sequences and highlight the existence of an underlying structure, often referred to as the "language of soccer". However, they do not incorporate player-specific information and are not designed to identify the actor of the next event. As a result, they fall short of explicitly modeling \textit{who} will perform the next action, even though they may predict \textit{what}, \textit{where}, and \textit{when} it will occur.

\subsection{Denoising Sequence-to-Sequence Pre-training for Natural Language Processing}

Self-supervised learning has proven effective for a wide range of NLP tasks, offering a framework for learning general-purpose representations without labeled data~\cite{surveyPTM}. Among these, denoising autoencoders have emerged as an effective approach for generative sequence modeling, wherein a model is trained to reconstruct original text from corrupted inputs~\cite{surveyPTM}. BART (Bidirectional and Auto-Regressive Transformers) \cite{bart} is a sequence-to-sequence model that combines a bidirectional encoder with an autoregressive decoder, trained as a denoising autoencoder. This task enables the model to learn rich representations that capture both global structure and local dependencies.

Drawing on the core idea behind BART, we treat context-free sequences of soccer action predictions as "corrupted inputs" and train a model to recover coherent and contextually plausible event sequences. While BART introduces noise to textual data through token masking, deletion, insertion, or reordering, we extend this principle to sequences of spatio-temporal action predictions produced by a model that can only attend frames within a short time window, and we enrich these inputs with game-state information. This framing allows us to view our model as learning the "language of soccer," where situated actions—like words—form structured sequences shaped by tactical and temporal dynamics. By training it to correct noisy, context-free predictions using broader game context, we enable the model to generate more consistent, player-specific, context-aware action sequences, analogous to how BART reconstructs fluent text from noised input.

\section{Methodology}
\subsection{Dataset}

We built and used two datasets: one for the STAD detector, and one for the "denoising" sequence transduction method.

\subsubsection{Footovision STAD Dataset}

This is an "on-the-ball events" dataset  composed of 20000 videos clips from 997 different and diverse football games, with a minimum of 2500 samples per class. Clips are at least 3 seconds long, and the frame rate is consistent at 25 frames per second.
There are 8 classes of action: ball-drive, pass, cross, header, throw-in, shot, tackle and ball-block. The sequences are sampled around randomly selected events, ensuring no overlap between clips. Each game was associated with either the training set or the test set, ensuring no contamination between them. We refer the reader to Ochin et al. \cite{ochin2025} for further information about this dataset.

\subsubsection{Full-length Game Footovision Dataset}

The dataset comprises 299 full-length game videos selected from a total of 997, each enriched with frame-by-frame tracking data, team formations, player positions, and simplified player-to-role assignments clustered into 13 categories (e.g., Goalkeeper, Full Back, Winger, Defensive Midfielder).

Additional metadata is provided, along with detailed frame-by-frame event annotations specifying both the action class and the identity of the player performing the action. The mapping between player identity and positional / tracking data is also included. The dataset splitting is consistent with one from the Footovision STAD Dataset.

All data is available for the full duration of each video. Tracking data is recorded continuously, including during game pauses, and some annotated events may occur during broadcast replays.

\subsection{Prior Predictions - Short time-window}

A lightweight implementation of the Track-Aware Action Detector (TAAD) is trained on the Footovision STAD dataset, as described in \cite{ochin2025}. The trained model is then applied to all 299 videos in the Full-length Game Footovision Dataset, using tracking data in the form of player bounding box sequences. Inference is performed by dividing each video into adjacent segments of T = 50 frames.

Missing bounding boxes—due to occlusion or tracking failure—are interpolated if the gap does not exceed three seconds. Otherwise, a fixed "dummy" bounding box is provided, along with a binary mask indicating its absence. This mask is used to zero out the corresponding ROI Align features before aggregation.

For a clip of T frames, with N players and K action classes, the output of TAAD is a tensor of shape $\mathbb{R}^{T \times N \times K}$, without softmax activation (the logits). To reduce border effects, inference is performed twice per video, with a $\frac{T}{2}$ frame offset between passes. The resulting tensors are temporally aligned and averaged.

While typical pipelines post-process these predictions before thresholding (e.g., with label smoothing \cite{ochin2025,Saha_2016,Singh_2017,kalogeiton_action_2017}), we opt to retain the raw logits. We hypothesize that this preserves richer information, allowing the model to learn patterns that might be lost through early discretization. Final tensors are stored without any post-processing.

\subsection{Role-based Game-State Representation}

Unlike standard STAD methods that predict actions independently for each player, we adopt an approach that reasons at the game level, across both teams, and outputs sequences of structured predictions as tuples (\textit{who}, \textit{what}, \textit{when}). To build this unified, structured representation of the full game state, it is necessary to address the player permutation problem~\cite{WeiFormation2013,Bialkowski2014}. For example, there are \(11!\) (almost 40 million) possible ways to order the positional features of 11 players, all encoding equivalent information about the team. We resolve this by enforcing a consistent, role-based concatenation order across all games.

For each team, players are ordered as follows: Goalkeeper, Left Back, Left Center Back, Mid Center Back, Right Center Back, Left Midfielder, Right Midfielder, Defensive Midfielder, Attacking Midfielder, Left Winger, Right Winger, Central Forward, and Right Back. Our game-state vector at each frame is constructed by concatenating individual player features—normalized position, velocity, and visibility—following this predefined role order, starting with the team attacking from the left (with pitch coordinates originating at the top-left, from the broadcast view).

To ensure fixed dimensionality despite substitutions, exclusions, or role mismatches (13 roles vs. 11 players), we represent the game-state for each frame as a vector in $\mathbb{R}^{26*D}$, where $D$ is the per-player feature dimension. Unoccupied roles are padded with a constant value of -15.0, well outside the valid feature range. This same role-based ordering is also applied to the unnormalized, context-free predictions (logits) generated by TAAD, as illustrated in Figure~\ref{fig:schema_global}.

\subsection{Denoising Sequence Transduction Model (DST)}

We implement a Transformer encoder-decoder architecture following the design proposed by Vaswani et al. \cite{Vaswani2017}, using 6 encoder and decoder layers, 8 attention heads, and a hidden dimension of 512. The encoder receives sequences of fixed length \(L = 750\) tokens (one token per frame). This sequence is constructed by concatenating the following components on the feature dimension:

\begin{center}
\(
\begin{array}{ll}
\text{Context-free predictions:} & \mathbb{R}^{L \times 234} \\
\text{Game-state vectors:} & \mathbb{R}^{L \times 130} \\
\text{Frame index (one-hot):} & \mathbb{R}^{L \times (L+2)} \\
\end{array}
\)
\end{center}

Here, the dimension 234 corresponds to 9 action classes across 26 roles, and 130 corresponds to 5 features per role (position \(x, y\), velocity \(v_x, v_y\), and visibility). The frame number encoding includes two additional tokens for start-of-sequence (SOS) and end-of-sequence (EOS), resulting in \(L+2\) dimensions. We use a sinusoidal positional encoding as described in~\cite{Vaswani2017}.

\bigskip

During training, the decoder receives target action sequences of variable length L'. This sequences of token is constructed by concatenating the following components on the feature dimension:

\begin{center}
\(
\begin{array}{ll}
\text{Action class (one-hot):} & \mathbb{R}^{L' \times 10} \\
\text{Player role (one-hot):} & \mathbb{R}^{L' \times 26} \\
\text{Frame index (one-hot):} & \mathbb{R}^{L' \times L+2} \\
\end{array}
\)
\end{center}

The action class vector includes 8 classes, plus 2 to encode the SOS and EOS tokens. The player role vector encodes 13 roles per team (26 total), and the frame index is one-hot encoded over the same temporal window with SOS and EOS.

The decoder attends to the encoder output via cross-attention and is trained with a causal mask to predict the next token in the sequence. Three cross-entropy loss functions are used—one for each component of the action prediction: class, role, and frame number. The SOS and EOS tokens are represented using special action class values and correspond, respectively, to frame 0 and frame \(L+1\).

At inference time, the "clean" action sequence is generated autoregressively: the input sequence is encoded once, and the decoder is called recursively, each time feeding it the previous prediction (class, role, frame), beginning with the SOS token and stopping when the EOS token is generated.

\subsection{Implementation details}

We used T = 50 frames as input for the TAAD, with no sub-sampling (2 seconds of video clip), and L = 750 frames for the transduction task. We trained the DST with the AdamW optimizer \cite{AdamW2019}, with a learning rate of 0.00025, a weight decay of 1e-05 on the non-bias parameters and a batch size of varying size (40 to 64) depending on the GPUs (RTX 3090 and RTX A6000 GPU) and experiments. We trained the system for a total of 6 epochs and divided the learning rate by 10 at epoch 3. For each epoch, 500 sequences of 750 frames were randomly sampled \textbf{in each} of the 240 full-length games of the training set.

\section{Evaluation}
\subsection{Metrics}

The predictions generated by the DST model are sequences of "discrete" predictions and do not require post-processing techniques such as label smoothing~\cite{ochin2025,Saha_2016,Singh_2017,kalogeiton_action_2017}. Thus, metrics are computed directly by thresholding the confidence scores. We report overall and per-class Precision and Recall at a low threshold of 15\%, reflecting the method’s intended use as a tool to assist annotation operators by proposing candidate actions. This setup prioritizes high recall to minimize missed detections, with the goal of achieving the best possible precision in that regime.

Evaluation is performed by matching predicted and ground-truth actions for the same player and class within a fixed temporal window (+/- $\delta$ around the annotation frame), with $\delta$ = 12 or 25 frames. A match is counted as a true positive (TP); unmatched predictions are false positives (FP), and unmatched ground-truth events are false negatives (FN). This matching procedure allows for the computation of Precision and Recall, both overall and per class.

\subsection{Comparisons}

To assess the benefit of reasoning at the game level, i.e. using role-based structured information, to denoise long sequences of detections in a high-recall, low-precision regime, we compare several setups and conduct an ablation study on the use of game-state information.

\subsubsection{Track-Aware Action Detector (TAAD):} We apply label smoothing post-processing~\cite{ochin2025,Saha_2016,Singh_2017,kalogeiton_action_2017} to the predictions generated by our baseline method. This model was trained on short clips of 50 frames containing only active gameplay and thus has no opportunity to learn when the game is paused (e.g., when the ball is out of play before a throw-in). In contrast, we expect our method to learn such contextual cues from game-state information. To ensure a fair comparison, we filter out detections that occur during these paused phases in the baseline predictions.

\subsubsection{Denoising Sequence Transduction (TAAD + DST):} We train two variants of the proposed DST model. The first uses both the prior predictions and the structured game-state representation; the second relies solely on the predictions, without access to player positions, velocities, or visibility. Both models are trained to denoise sequences of 750 frames. The results of this comparison are presented in Table~\ref{tab:PrecRecComp}.

\begin{table}[h!]
\centering
\caption{Comparison of the Precision (\textbf{PR}) and Recall (\textbf{REC}) with a confidence score threshold of 15\% and $\delta$ = 12 frames, between TAAD~\cite{singh2022} and 2 variants of our method, with and without game-state information. Results are consistent with $\delta$ = 25. Metrics are $\times 10^2$.}
\resizebox{\textwidth}{!}{
\begin{tabular}{MM|N|N||N|N|N|N}
\hline
\textbf{Class} & \textbf{Samples} & \multicolumn{2}{c||}{\textbf{TAAD \cite{singh2022}}} & \multicolumn{2}{c|}{\textbf{TAAD + DST}} & \multicolumn{2}{c}{\textbf{TAAD + DST}} \\ 
& & \multicolumn{2}{c||}{} & \multicolumn{2}{c|}{\small{without Game State}} & \multicolumn{2}{c}{\small{with Game State}} \\ 
\cline{3-8}
 &  & \textbf{PR} & \textbf{REC} & \textbf{PR} & \textbf{REC} & \textbf{PR} & \textbf{REC} \\ \hline
Pass & 26,428 & 62.6 & 71.1 & 79.1 & 74.5 & \textbf{83.0} & \textbf{79.8} \\ 
Ball-drive & 20,772 & 37.5 & 68.0 & 74.0 & 71.6 & \textbf{79.6} & \textbf{77.5} \\ 
Header & 2,232 & 26.9 & 49.4 & 44.9 & 44.1 & \textbf{46.3} & \textbf{45.9} \\ 
Cross & 1,291 & 60.2 & 55.8 & 65.2 & 65.5 & \textbf{67.4} & \textbf{67.2} \\ 
Throw-in & 1,149 & 5.8 & 24.2 & 51.7 & 47.8 & \textbf{68.5} & \textbf{65.3} \\ 
Ball-block & 764 & 10.5 & 44.5 & 35.8 & 22.3 & \textbf{38.2} & \textbf{29.6} \\ 
Shot & 669 & 47.8 & 59.3 & 66.5 & 59.9 & \textbf{69.7} & \textbf{63.5} \\ 
Tackle & 134 & 3.0 & 6.0 & \textbf{9.3} & \textbf{3.0} & 7.0 & 2.2 \\ \hline
\textbf{Overall} & 53,439 & 43.5 & 66.9 & 74.1 & 70.2 & \textbf{78.7} & \textbf{75.8} \\ \hline
\end{tabular}
}
\label{tab:PrecRecComp}
\end{table}

Additionally, to measure the influence of the temporal context on our method, we train 3 variants with different temporal coverage: 100, 250 and 750 frames. The results are presented in Table~\ref{tab:temporalcontext}.

\begin{table}[h!]\centering
\caption{Influence of the temporal context on Precision (\textbf{PR}) and Recall (\textbf{REC}), all other things being equals. Metrics are $\times 10^2$.}
\includegraphics[width=.99\linewidth]{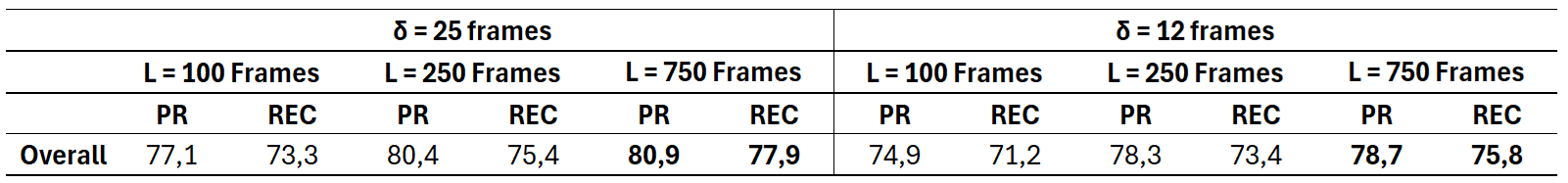}
\label{tab:temporalcontext}
\end{table}

\subsection{Discussion}

The results in Table~\ref{tab:PrecRecComp} demonstrate the separate contributions of both role-based structured information and game-state information.

First, we observe a clear benefit of reasoning at the game level: the absence of multiple simultaneous detections across players indicates that this approach naturally enforces the constraint that only one player can interact with the ball at any given time. More broadly, both Precision and Recall improve when transitioning from context-free, player-centric predictions to unified predictions based on a structured representation of the same underlying data. The ablation study further emphasizes the important role of game-state information in this improvement. We analyze the specific contributions to Precision and Recall in more detail below.

\subsubsection{Gains in Precision:} We first notice that most of the improvement in Precision is achieved by transitioning from per-player predictions made by TAAD to game-level predictions using a role-based structured representation of the initial outputs, without any game-state information (i.e., no position, velocity, or visibility features). We attribute this to two main factors. First, the roles assigned to players implicitly encode stable relative positional information between players, making it likely that certain patterns (e.g., passes) emerge between specific roles, aligned with team strategies during different phases of play (e.g., build-up, progression, final third, offensive/defensive transitions). Second, game-level prediction encourages the model to select the most plausible action based on broader temporal context, even in cases where multiple visually ambiguous false positives would otherwise be produced.

The addition of game-state information yields a further 4.5\% improvement in Precision, consistent with observations from previous work~\cite{ochin2025}.

\subsubsection{Gains in Recall:} Given the extremely low confidence threshold used and the significant increase in Recall, we conclude that the method enables the recovery of "invisible" events that can only be inferred from context, rather than from visual features alone. Our ablation analysis suggests that approximately two-thirds of this gain originates from the inclusion of game-state information, while the remaining one-third is due to switching to a game-level prediction based on role-based representations.

\subsubsection{Influence of Temporal Coverage:} As expected, both Precision and Recall benefit from a larger temporal context, with diminishing returns for Precision beyond 250 frames, suggesting a potential performance plateau. Further experiments are needed to determine whether Recall also plateaus beyond 750 frames. Considering that a typical phase of play in soccer lasts 2–3 minutes, it would be interesting to evaluate our method on longer sequences (e.g., 1500 or 3000 frames).
We also observe substantial improvement even with only 100 frames of context, on average containing just two actions. A likely explanation lies in the Markovian hypothesis, widely used in soccer analytics: the next action in a game depends primarily on the current state and/or the most recent action. This aligns with the intuition that short-term cues, such as a player’s immediate movement, ball proximity, or defensive pressure, carry significant predictive power. Many actions in soccer, like passes or tackles, are reactive and highly localized in time, meaning even limited context often includes the key triggers that precede an event.

\subsubsection{Discrepancy in Performance Across Classes:} We observe that Precision and Recall for the \textit{Header}, \textit{Ball-block}, and \textit{Tackle} classes remain significantly lower than for other classes. These three classes are also those for which tracking data is frequently missing at the moment of the action, likely due to frequent occlusions or close proximity between players. Moreover, \textit{Ball-block} and \textit{Tackle} are two minority classes in our dataset, accounting for only 1.6\% of all events. Developing a more elaborate method than bounding box interpolation may improve this issue and is left for future work.

\section{Conclusion}

This work presents a novel approach to spatio-temporal action detection in soccer, recasting the task as a denoising sequence transduction problem over long sequences of noisy, context-free player-centric action predictions. By structuring these predictions through role-based concatenation and incorporating game-state features such as player positions, velocities, and visibility, we enable a transformer-based model to recover coherent, tactically plausible, context-aware sequences of actions. The results demonstrate substantial gains in both precision and recall, particularly in the high-recall regime, which is critical for supporting human annotators.
Our DST model significantly enhances the performance of TAAD, demonstrating its potential as an effective assistant for annotation operators. Its ability to improve detections from a noisy and relatively low-dimensional input signal naturally raises the question of how much further performance could be improved using a richer and more task-agnostic representations. As a promising direction for future work, we plan to explore the integration of generic visual features extracted from off-the-shelf vision foundation models, potentially fine-tuned in a self-supervised manner on our dataset. Such representations could capture richer, higher-level semantics while remaining compatible with our current training setup, which does not require end-to-end optimization of the feature extractor.

%
%
%
%

\end{document}